# FSampler: Training-Free Acceleration of Diffusion Sampling via Epsilon Extrapolation

Author: Michael A. Vladimir

## Abstract

FSampler is a training-free, sampler-agnostic execution layer that accelerates diffusion sampling by reducing the number of function evaluations (NFE). FSampler maintains a short history of denoising signals (epsilon) from recent real model calls and extrapolates the next epsilon using finite-difference predictors at second order (h2), third order (h3, Richardson), or fourth order (h4), falling back to lower order when history is insufficient. On selected steps the predicted epsilon substitutes the model call while keeping each sampler's update rule unchanged. Predicted epsilons are validated for finiteness and magnitude; a learning stabilizer rescales predictions on skipped steps to correct drift, and an optional gradient-estimation stabilizer compensates local curvature. Protected head/tail windows, periodic anchors, and a cap on consecutive skips-bound deviation over the trajectory. Operating at the sampler level, FSampler integrates with Euler/DDIM, DPM++ 2M/2S, LMS/AB2, and RES-family exponential multistep methods and drops into standard ComfyUI workflows.

Across FLUX.1-dev, Qwen-Image, and Wan 2.2, FSampler reduces wall-clock time by ~8-22% and model calls by ~15-25% at high fidelity (Structural Similarity Index (SSIM) ~ 0.95-0.99), without altering sampler formulas. With an aggressive adaptive gate, reductions can reach ~45-50% fewer model calls at lower fidelity (SSIM ~ 0.73-0.74).

These image changes are often imperceptible on first glance when compared to baseline (see Section 4). FSampler composes with model-level optimizations (e.g., distillation, quantization) and scheduler choices, enabling practical, training-free acceleration in real workflows. Community testers have confirmed success on SDXL, Chroma and more. Code available at: [ComfyUI-FSampler](#). Experiment data (including high-resolution images) available at: [FSampler_ExperimentData](#). ComfyUI is an open-source, node-based diffusion workflow graphical user interface (GUI); get it from the [ComfyUI GitHub](#) or the [official site](#) for the standalone installer.

## 1. Introduction

Diffusion sampling remains computationally expensive because high-quality solvers require many model calls. Practical workflows benefit from training-free acceleration that preserves perceptual fidelity and drops into existing pipelines without modifying models or sampler implementations. FSampler targets this need with a simple, sampler-agnostic layer that reduces the number of model calls while maintaining output quality.

FSampler predicts the denoising signal from recent real steps and substitutes the predicted denoised value on selected steps, leaving each sampler's update rule unchanged. Before a skip is accepted, the predicted signal is validated for numerical sanity and reasonable magnitude. Two stabilizers keep trajectories on track: a learning stabilizer that corrects systematic scale drift, and an optional gradient-estimation stabilizer that compensates local curvature on skip steps. Guard rails bound drift over the trajectory, including protected head and tail windows, periodic anchors that force REAL calls, and a cap on consecutive skips. Because FSampler operates at the sampler execution level, it plugs into common families-Euler/DDIM, DPM++ 2M/2S and LMS (AB2), and RES-style exponential multistep methods-without altering their formulas.

Across modern text-to-image models, representative configurations achieve consistent efficiency gains at high fidelity. On FLUX.1-dev, a conservative setting (h2/s4 with the learning stabilizer) reaches SSIM=0.9818 with 15.9% wall-clock time saved; a balanced setting (h2/s3 with the learning stabilizer) attains SSIM=0.9533 with 21.6% time saved. Aggregating across models, configurations with SSIM $>= 0.95$ typically realize ~8-22% time savings and ~15-25% fewer model calls. In practice, SSIM $>= 0.90$ often corresponds to outputs that are different rather than degraded: realism and style adherence are preserved while following an alternate but plausible denoising trajectory (see Section 4).

Adaptive skipping is available for users who prefer dynamic control. The anchor interval forces periodic real calls, the maximum number of consecutive skips prevents long runs without ground-truth correction, and explicit skip indices let users choose the exact steps to skip. With per-model tuning, adaptive skipping is especially useful for quick previews of likely final images and fast seed sweeps before committing conservative skip calls; aggressive acceptance rates can reduce perceptual similarity, as discussed in the limitations.

FSampler is orthogonal to model-level optimizations and schedulers and is compatible with quantized or distilled models that retain standard denoised/epsilon interfaces. Preliminary use in video pipelines indicates compatibility, though systematic evaluation is needed to quantify temporal coherence and frame-to-frame consistency. To accelerate adoption and validation in real workflows, FSampler is released as a drop-in ComfyUI node with diagnostics and experiment logging. This choice enables immediate integration into existing graphs and rapid community feedback across diverse setups-particularly valuable for an independent author without access to multiple facilities.

The paper proceeds as follows: Section 2 introduces the background and notation. Section 3 describes the method, including predictors, policies, validation, stabilizers, and sampler integrations. Section 4 presents results and ablations across models and samplers. Section 5 concludes with limitations and outlook.

## 2. Background

This section defines the minimum concepts needed to follow FSampler in plain terms: what denoised, epsilon, derivative, time, and log_snr mean; the sampler updates they feed; and how NFE is counted. It also shows where FSampler intervenes: on a skip step, the sampler uses a predicted denoised/epsilon instead of calling the model; the sampler's update remains unchanged.

**Diffusion ODE and notation.** Diffusion sampling is framed as integrating the probability flow ordinary differential equation (ODE) for the latent state under a noise-scale schedule. The core variables used throughout this paper are:

```
# State and noise schedule
x: latent state at current step
sigma: noise scale at current step
log_snr = -log(sigma)  # log-SNR coordinate used by exponential integrators

# Model predictions
denoised = model(x, sigma)  # predicted clean image (x0)
epsilon = denoised - x      # noise residual
derivative = (x - denoised) / sigma   # ODE derivative, equivalently -epsilon/sigma

# Step progression
sigma_current: noise scale at step n
sigma_next: noise scale at step n+1
log_snr_step = -log(sigma_next) - (-log(sigma_current))  # step size in log-SNR space
```

The probability flow ODE can be expressed in two equivalent forms:

```
# Time-based formulation (use explicit name "time")
dx/dtime = f(x, time)

# Log-SNR formulation (used by RES, DEIS)
dx/dlog_snr = g(x, log_snr)    where log_snr = -log(sigma)
```

Samplers advance the state from step to step by calling the model to obtain denoised, then applying an integration rule.

**Discretizations and samplers.** First-order schemes such as Euler and DDIM (Song et al., 2021) advance the state using the current derivative:

```
# Euler/DDIM update
denoised = model(x, sigma_current)
derivative = (x - denoised) / sigma_current
time = sigma_next - sigma_current
x_next = x + derivative * time
```

DDIM equivalently interpolates in denoised space with the same notion of time.

Second-order linear multistep methods (LMS, DPM++ 2M) employ the Adams-Bashforth AB2 formula utilizing two consecutive derivative evaluations:

```
# DPM++ 2M / LMS (AB2) update
denoised_current = model(x, sigma_current)
derivative = (x - denoised_current) / sigma_current
derivative_previous = ... # stored from previous step

time = sigma_next - sigma_current
x_next = x + time * (1.5 * derivative - 0.5 * derivative_previous)
```

Exponential integrators (DEIS, RES-family) operate in log-SNR space with $\phi$-functions yielding multistep coefficients:

```
# RES-2M exponential multistep update
denoised_current = model(x, sigma_current)
epsilon_current = denoised_current - x
epsilon_previous = ... # stored from previous step

log_snr_current = -log(sigma_current)
log_snr_next = -log(sigma_next)
log_snr_step = log_snr_next - log_snr_current

# Coefficients from ϕ-functions (coeff1, coeff2 computed by the sampler)
x_next = x + log_snr_step * (coeff1 * epsilon_current + coeff2 * epsilon_previous)
```

These families cover the standard discretizations used in modern text-to-image systems and form the integration points where FSampler substitutes predicted denoised values on selected steps without altering the underlying update rules. These families align with established solvers such as DPM-Solver (Lu et al., 2022a), DPM++ 2M (Lu et al., 2022b), UniPC (Zhao et al., 2023), DEIS (Zhang and Chen, 2022), and RES multistep methods (Zhang et al., 2023).

**Schedules and NFE.** A scheduler defines the sequence of noise scales across the denoising trajectory:

```
# Scheduler output
sigma_schedule = [sigma_0, sigma_1, sigma_2, ..., sigma_N]
# Equivalently in log-SNR space
log_snr_schedule = [-log(sigma_0), -log(sigma_1), ..., -log(sigma_N)]

# Step size at each transition
log_snr_step_n = log_snr_schedule[n+1] - log_snr_schedule[n]
```

Common schedules include uniform steps in log-SNR space (simple scheduler) and two-stage schedules that apply different spacing regimes for high-noise and low-noise regions (e.g., beta followed by bong_tangent). Computational cost is reported as the Number of Function Evaluations (NFE), defined as the total number of model forward passes across the trajectory: sum all model calls performed at each REAL step, and count 0 for SKIP steps. Wall-clock time is measured end-to-end for the complete sampling trajectory. Deterministic zero-noise schedules yield identical outputs for fixed random seeds; when stochastic noise is enabled, trajectories remain seed-deterministic in this implementation.

**Notation map (to conventional symbols).** The paper uses descriptive names to improve accessibility; the following map provides one-to-one links to common symbols used in prior work for cross-reference:

```
noise_scale <-> σ
log_snr <-> λ = -log σ
time <-> σ_next - σ_current (difference between consecutive noise levels)
denoised <-> x0 or x0_hat
epsilon <-> denoised - x (often written ε)
derivative <-> (x - denoised) / noise_scale = -epsilon / noise_scale
log_snr_step <-> λ_{n+1} - λ_n
coeff1, coeff2 <-> b1, b2 (φ-function-based weights in exponential multistep methods)
```

# 3. Method

## 3.1 Epsilon Extrapolation Formulas

FSampler maintains a history of epsilon predictions from recent model evaluations and uses finite-difference extrapolation to predict the next epsilon without calling the model. Given epsilon history from prior real steps indexed by n, three predictor orders are defined based on the number of available history points.

**Second-order linear extrapolation (h2)** uses two prior evaluations:

```
epsilon_hat = 2 * epsilon[n-1] - epsilon[n-2]
```

**Third-order Richardson extrapolation (h3)** requires three history points:

```
epsilon_hat = 3 * epsilon[n-1] - 3 * epsilon[n-2] + epsilon[n-3]
```

**Fourth-order cubic extrapolation (h4)** uses four history points:

```
epsilon_hat = 4 * epsilon[n-1] - 6 * epsilon[n-2] + 4 * epsilon[n-3] - epsilon[n-4]
```

When insufficient history is available, FSampler falls back to the next lower order following the ladder h4 -> h3 -> h2. Once epsilon_hat is computed, the sampler receives the predicted denoised value in place of calling the model. On real steps where the model is called, the true epsilon is computed and appended to history:

```
    # SKIP step
    denoised = x + epsilon_hat

    # REAL step
    denoised = model(x, sigma)
    epsilon = denoised - x
    epsilon_history.append(epsilon)
```

Implementation: sampling/extrapolation.py:4-50.

## 3.2 Skip Policies

FSampler supports two skip policies: fixed patterns and adaptive gating.

**Fixed patterns (hN/sK).** Fixed patterns follow a deterministic cadence where the model is called K times followed by a single skip, yielding a cycle length of K+1. Patterns are denoted as hN/sK where N is the predictor order (h2, h3, or h4) and K is the number of real calls before each skip.

Common patterns include h2/s2 which executes Call, Call, Skip with cycle length 3 for approximately 33% NFE reduction, h3/s3 which uses Call, Call, Call, Skip with cycle length 4 for approximately 25% reduction, and h4/s4 which performs four model calls followed by one skip for approximately 20% reduction.

To prevent degradation, fixed patterns protect the first protect_first_steps and last protect_last_steps of the sampling trajectory. The pattern activates only after sufficient history is available, specifically max(protect_first_steps, history_order). The skip decision first verifies that step_index >= protect_first_steps and step_index < total_steps - protect_last_steps, then confirms len(epsilon_history) >= history_order. Given these constraints, the algorithm computes anchor = max(protect_first_steps, history_order), cycle_length = skip_calls + 1, and cycle_position = (step_index - anchor) mod cycle_length, accepting the skip when cycle_position equals cycle_length - 1. Implementation: sampling/skip.py:124-228.

**Adaptive gate.** The adaptive gate compares two predictors of different orders to estimate local error and accepts the skip if the discrepancy falls below a tolerance threshold. Specifically, epsilon_hat_high is computed using third-order Richardson extrapolation (h3) and epsilon_hat_low using second-order linear extrapolation (h2), then the relative error is computed as

```
    relative_error = RMS(epsilon_hat_high - epsilon_hat_low) / max(RMS(epsilon_hat_high), 1e-6)
```

where RMS(tensor) = sqrt(mean(tensor**2)). If relative_error <= tolerance, the skip is accepted and epsilon_hat_high is used; otherwise, a real model call is performed.

Adaptive mode employs the same protected head and tail windows as fixed patterns, but adds two additional safeguards to prevent drift. An anchor interval forces a real call every Nth step regardless of error estimates, ensuring periodic ground truth updates. A max consecutive skips limit caps back-to-back skips to maintain regular model evaluations. The adaptive gate requires a minimum of 3 real epsilons in history for dual-predictor comparison.

When sampler state is available (x, sigma_current, sigma_next, sampler_kind), FSampler compares predicted next states in latent space rather than epsilon space, providing more robust error estimates for complex samplers like DPM++ 2M. Implementation: sampling/skip.py:263-401.

**Explicit skip indices (override).** For targeted experiments, users may specify explicit step indices to skip (e.g., "h3, 6, 9, 12"). The first token hN selects the predictor (defaults to h2 if omitted). Indices are 0-based after any start/end slicing; 0 and 1 are never skipped, and indices are bounded to the valid range. When provided, explicit indices override fixed/adaptive gating and associated guard rails (anchors, max-consecutive, protected windows).

The sampler still requires sufficient REAL epsilon history for the chosen predictor (falls back to lower order when needed) and applies the learning/clamp behavior described below. Implementation: sampling/skip.py:72 (parsing) and per-sampler explicit branches.

## 3.3 Validation and Stabilizers

Before accepting a predicted epsilon_hat for a skip step, FSampler applies validation checks to detect unstable extrapolations that could degrade sample quality.

**Validation procedure.** Before accepting a predicted epsilon_hat for a skip step, the sampler applies a single, shared validation procedure: (1) reject any epsilon_hat containing NaN/Inf or with non-finite norm; (2) enforce an absolute floor ||epsilon_hat|| >= 1e-8; (3) when a previous real epsilon is available, enforce a relative floor ||epsilon_hat|| >= 1e-6 · ||epsilon_prev||. If any check fails, the skip is cancelled and a real model call is performed instead. Implementation: sampling/skip.py:231-260.

RES-family additional guard. On top of the validation procedure, RES-family samplers (RES-2M and RES-multistep) apply an extra magnitude cap that cancels a skip when the prediction is excessively large relative to the last real epsilon: ||epsilon_hat|| > 50 · ||epsilon_prev||. This "too_large_rel" guard complements the validation procedure and is specific to the RES family (Euler/DDIM/AB2 rely on the validation procedure alone, plus any learning/grad-est clamps). Implementation: sampling/samplers/res2m.py:61,120; sampling/samplers/res_multistep.py:508.

**Learning stabilizer (learning_ratio).** To correct systematic over- or under-prediction in the extrapolation formulas, FSampler employs an exponential moving average (EMA) learning ratio, referred to here as learning_ratio. After each real step where both epsilon_hat and epsilon_real are available, the observation ratio is computed as

```
learn_observation = ||epsilon_hat|| / (||epsilon_real|| + 1e-8)
```

The learning ratio is then updated via

```
learning_ratio = beta * learning_ratio + (1 - beta) * learn_observation
```

where beta is a smoothing factor; in our experiments we use β=0.9985 on FLUX.1-dev and β=0.995 on Qwen-Image and Wan 2.2. The learning_ratio is clamped to [0.5, 2.0] to prevent extreme corrections. On subsequent skip steps, epsilon_hat is scaled by 1/learning_ratio before use, i.e., epsilon_hat := epsilon_hat / learning_ratio. This scaling quietly corrects bias in the extrapolation without modifying the predictor formulas themselves. Implementation: sampling/learning.py:1-28.

**Gradient estimation correction.** For samplers that benefit from second-order accuracy, FSampler optionally applies a gradient estimation correction on skip steps. Given the ODE derivative derivative_hat = -epsilon_hat / sigma_current and the previous real derivative derivative_previous from the last model call, the correction term is computed as

```
derivative_correction = (curvature_scale - 1) * (derivative_hat - derivative_previous)
```

where curvature_scale defaults to 2.0. To maintain stability, the correction magnitude is clamped such that ||derivative_correction|| / (||derivative_hat|| + 1e-8) <= 0.25. The final update becomes x := x + (derivative_hat + derivative_correction) * time, where time = sigma_next - sigma_current. This approximates the local curvature of the trajectory without requiring additional model evaluations. The gradient estimation mode is activated via the grad_est or learn+grad_est adaptive modes. Implementation: sampling/samplers/gradient_estimation.py:1-329.

## 3.4 Integration with Samplers

FSampler integrates with existing samplers by substituting the model call with a predicted denoised value on skip steps, while preserving each sampler's characteristic update formula.

**Euler-like samplers (Euler, RES-2S, DPM++ 2S).** On skip steps, these samplers compute denoised = x + epsilon_hat, then form the ODE derivative derivative = (x - denoised) / sigma_current and apply the first-order update x := x + derivative * time where time = sigma_next - sigma_current. When gradient estimation is enabled, derivative_correction is added to derivative before the update. Implementation: sampling/samplers/euler.py, res2s.py, dpmpp_2s.py.

**DDIM (Song et al., 2021).** DDIM computes the predicted clean image x0_hat = x + epsilon_hat, then applies its characteristic interpolation as a function of time between noise levels: x := x0_hat + scale * (x - x0_hat) where scale = sigma_next / sigma_current. This preserves DDIM's noise-level interpolation structure while skipping the model call. Implementation: sampling/samplers/ddim.py:36-38.

**Multistep Adams-Bashforth (DPM++ 2M, LMS) (Lu et al., 2022b).** These samplers use a second-order Adams-Bashforth formula. On skip steps, they compute derivative = -epsilon_hat / sigma_current and apply x := x + time *(1.5 derivative - 0.5 derivative_previous) when the previous derivative derivative_previous is available, falling back to x := x + time* derivative otherwise. The coefficients 1.5 and -0.5 are the standard AB2 weights and remain unchanged. Implementation: sampling/samplers/dpmpp_2m.py:48-53, lms.py.

**RES-2M (Zhang et al., 2023).** RES-2M is a second-order exponential multistep integrator in log_snr = -log σ with coefficients derived from φ-functions. On skip steps, epsilon_hat is substituted to form denoised = x + epsilon_hat, and the update uses the exponential multistep form

```
x_pre = x + log_snr_step * (coeff1 * epsilon_current + coeff2 * epsilon_previous)
```

where log_snr_step = (-log sigma_next) - (-log sigma_current), and (coeff1, coeff2) come from φ1/φ2 (with a geometry factor often denoted c2); if coefficients become invalid, an Euler fallback is used for the step. In learning mode (on real steps), RES-2M softly rescales (coeff1, coeff2) with a sum-preserving adjustment based on a smoothed epsilon-norm ratio; on SKIP steps, epsilon_hat is scaled by 1/learning_ratio as in Section 3.3.

**RES-multistep (general).** For RES-style multistep variants, on SKIP steps FSampler substitutes denoised = x + epsilon_hat (or epsilon_hat/learning_ratio in learning modes) within the multistep formula, then proceeds with the standard update. When enabled, a small post-integrator slope correction may be applied.

# 4. Experiments

## 4.1 Experimental Setup

FSampler is evaluated across three distinct text-to-image diffusion models to assess generalization across samplers, schedulers, and step counts. The experimental suite comprises 105 total runs (3 baselines + 102 FSampler configurations) spanning a broad matrix of skip patterns and adaptive modes (coverage varies slightly by model).

**FLUX.1-dev.** The first experimental suite uses FLUX.1-dev (Black Forest Labs, 2024) with the `res_2s` sampler and a `simple` scheduler. The baseline performs 20 sampling steps with 20 function evaluations (NFE=20) and completes in 160.26 seconds. This suite includes 42 experiments: 1 baseline and 41 FSampler configurations.

**Qwen-Image.** The second suite employs Qwen-Image (Qwen Team, 2024) with the `euler` sampler and a `simple` scheduler. The baseline executes 25 steps (NFE=25; 25 model calls across 50 schedule transitions) and completes in 145.39 seconds. This suite comprises 31 experiments: 1 baseline and 30 FSampler configurations.

**Wan 2.2.** The third suite tests a two-stage Wan 2.2 image pipeline (Wan Team, 2024) with the `res_2s` sampler in both stages and a combined `beta+bong_tangent` scheduler (high-noise stage then low-noise stage). The baseline performs 26 steps (NFE=26) in 213.59 seconds. This suite includes 32 experiments: 1 baseline and 31 FSampler configurations. Note: the stored identifier `wan22-high-noise+wan22-high-noise` reflects a naming constraint only; the second stage is low-noise in practice.

**FSampler configurations.** Each model is tested against a matrix of skip patterns and adaptive modes. Skip patterns include fixed-cadence modes `h2/s2`, `h2/s3`, `h2/s4`, `h2/s5` (second-order), `h3/s3`, `h3/s4`, `h3/s5` (third-order), and `h4/s4`, `h4/s5` (fourth-order), plus an `adaptive` gate mode that decides skips via dual-predictor error estimation. Adaptive modes used include `learning` (EMA-based scaling), `grad_est` (gradient estimation on skip steps), and `learn+grad_est` (both). The plain baseline corresponds to `skip_mode=none`. Guard rails are consistent across models with `max_consecutive_skips=2`; `anchor_interval=4` is standard. Protected head/tail windows vary slightly by workflow (commonly 0-1 head step and 1 tail step recorded in metadata), and do not affect the comparative conclusions.

**Evaluation metrics.** Quality is measured via Structural Similarity Index (SSIM), Root Mean Square Error (RMSE), and Mean Absolute Error (MAE) comparing FSampler outputs to same-seed baseline outputs. Efficiency is quantified by NFE (Number of Function Evaluations) reduction percentage-defined as the percent decrease in total model forward passes relative to the baseline-and wall-clock time saved percentage. All experiments use fixed random seeds per model to ensure deterministic comparison.

**Hardware and implementation.** All experiments run on identical hardware with the same ComfyUI workflow environment. Each configuration is executed once to measure runtime and quality metrics. The experimental data, analysis scripts, and generated images are available in the experiments/ directory organized by model identifier (fluxr1, qwen1, wan221), with automated analysis reports generated via analyze_experiments.py.

## 4.2 Main Results

The following analysis focuses on FLUX.1-dev results to demonstrate FSampler's quality-efficiency tradeoff in detail. FSampler reduces NFE by 15-25% on FLUX.1-dev with SSIM >= 0.95 (Figure 4.2a-c). The baseline performs 20 calls (NFE=20) in 160.26 s.

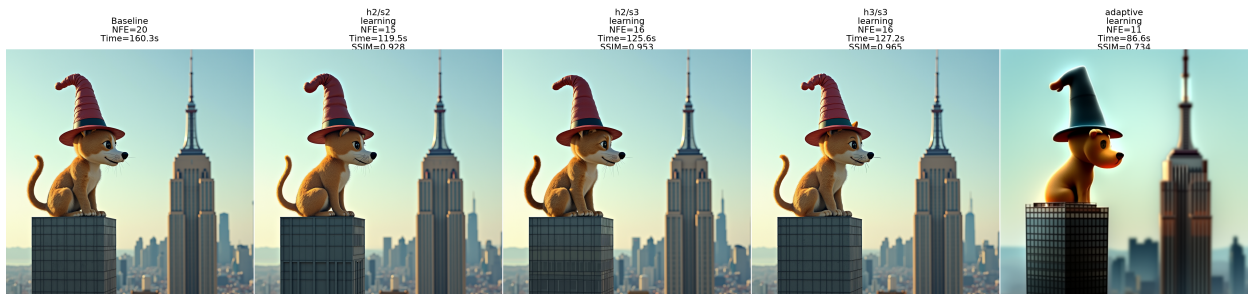

*Figure 4.2a: FLUX.1-dev curated strip (seed 2028). Baseline, h2/s2+L, h2/s3+L, h3/s3+L, and adaptive+L. Configurations with SSIM >= 0.95 are visually indistinguishable from baseline.*

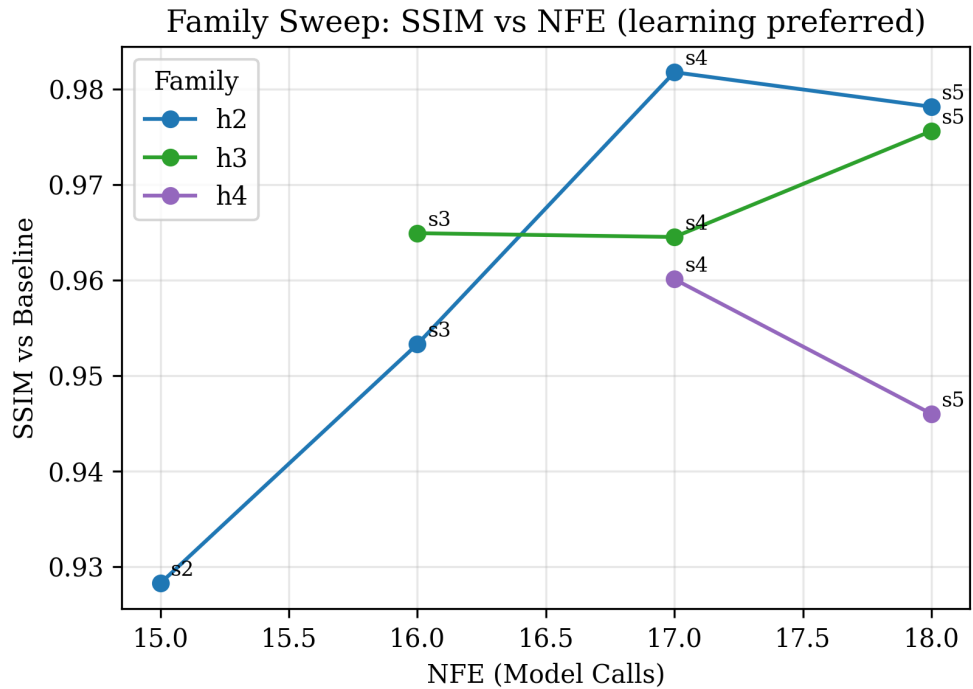

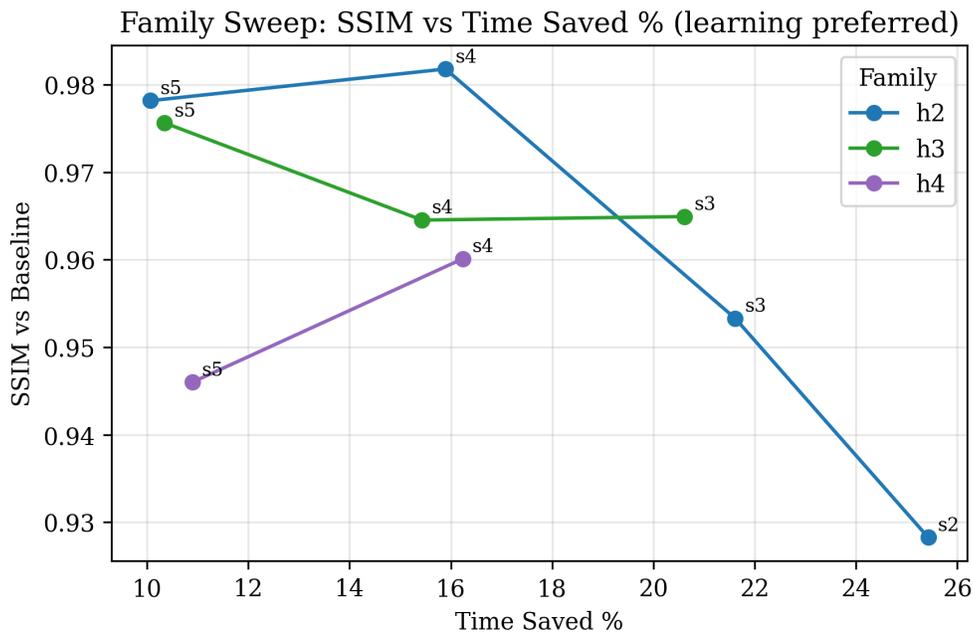

**Second-order h2 patterns dominate the quality-efficiency frontier on FLUX.1-dev** (Figures 4.2b-c). Conservative h2/s4+learning achieves SSIM=0.9818 with 15.0% NFE reduction (17/20 calls) and 15.9% time saved (134.79 s). Moderate h2/s3+learning reaches SSIM=0.9533 with 20.0% NFE reduction (16/20) and 21.6% time saved (125.62 s). Higher-order h3/h4 patterns show no consistent advantage on this sampler. Figure 4.2a confirms that SSIM >= 0.95 configurations are near-indistinguishable from baseline with only subtle differences; aggressive adaptive gating (SSIM=0.73) shows visible degradation.

## 4.3 Ablations

Figure 4.3 decomposes FSampler's quality and efficiency by skip pattern and adaptive mode on FLUX.1-dev.

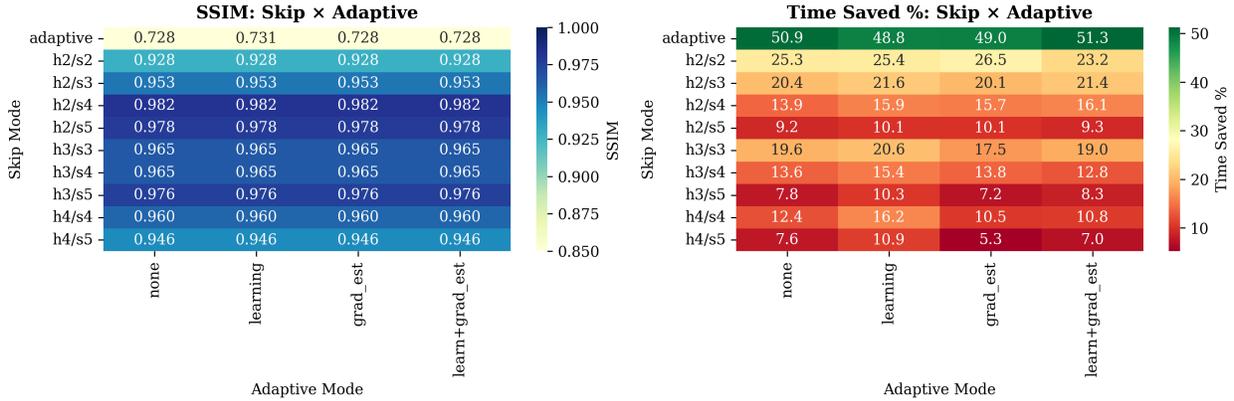

*Figure 4.3 (FLUX.1-dev): Ablation heatmaps. Left: SSIM by skip pattern x adaptive mode. Right: Time saved (%) by skip pattern x adaptive mode. Conservative h2/s3-s4 with learning stabilizer delivers the best balance.*

**Skip patterns.** Second-order h2 patterns form the quality-efficiency frontier. At 20% NFE reduction, h2/s3+learning achieves SSIM=0.9533 with 21.6% time saved; at 15% reduction, h2/s4+learning reaches SSIM=0.9818 with 15.9% time saved. At the same 20% NFE reduction, h3/s3+learning offers slightly higher SSIM (0.9649 vs 0.9533 for h2/s3+learning) but saves a bit less time (~20.6% vs 21.6%). h4/s4+learning is lower still (SSIM=0.960 at 15% reduction). Overall, h2 patterns define the most consistent quality-efficiency frontier on this sampler. The adaptive gate achieves 45-50% NFE reduction but reduces quality to SSIM~0.73-0.74, indicating that this aggressive skipping level is outside the stability range for this sampler.

**Adaptive modes.** Holding skip pattern constant at h2/s3 (20% NFE reduction), all four adaptive modes-learning, grad_est, learn+grad_est, none-produce identical SSIM=0.9533, RMSE=0.0354, and MAE=0.0135, indicating that periodic anchors (anchor_interval) maintain quality. Wall-clock efficiency differs: learning achieves 21.6% time saved (125.62 s), grad_est 20.1% (127.98 s), and none 20.4% (127.61 s). The learn+grad_est hybrid (21.4% saved, 126.02 s) shows no benefit over learning alone; gradient-estimation overhead does not translate into quality gains here.

Across these stable samplers, the learning stabilizer (L) has a subtle effect; it primarily serves to steer the trajectory back if it begins to drift. Its contribution is robustness rather than large headline gains.

## 4.4 Generalization Across Models

FSampler's fixed-cadence skipping with a learning stabilizer (EMA-based scaling) generalizes across different diffusion architectures, samplers, and schedulers.

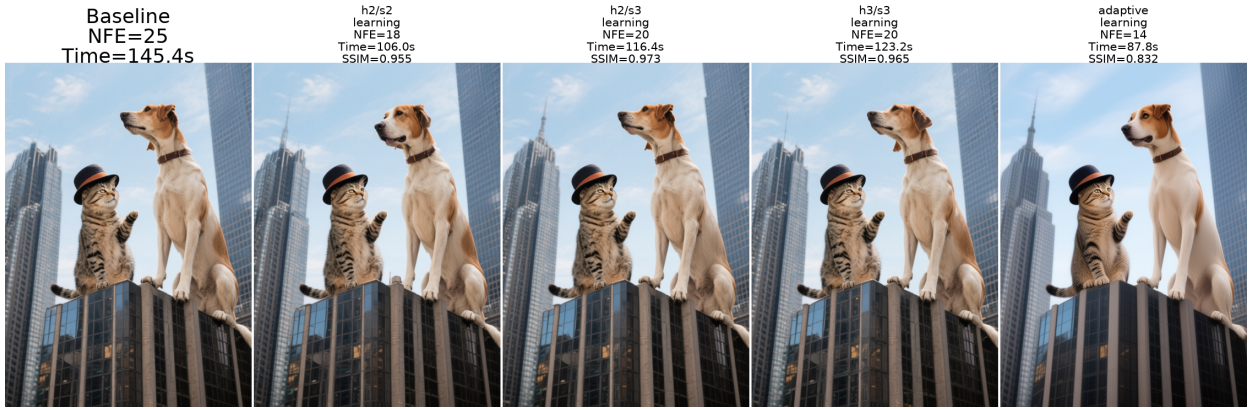

*Figure 4.4a: Qwen-Image visual comparison. euler sampler, simple scheduler, 25-step baseline (NFE=25; 25 calls across 50 transitions). Best: h2/s5+learning (best by SSIM; SSIM=0.9952, 8.1% time saved).*

Figures 4.4a show visual comparisons for Qwen-Image (euler sampler, simple scheduler).

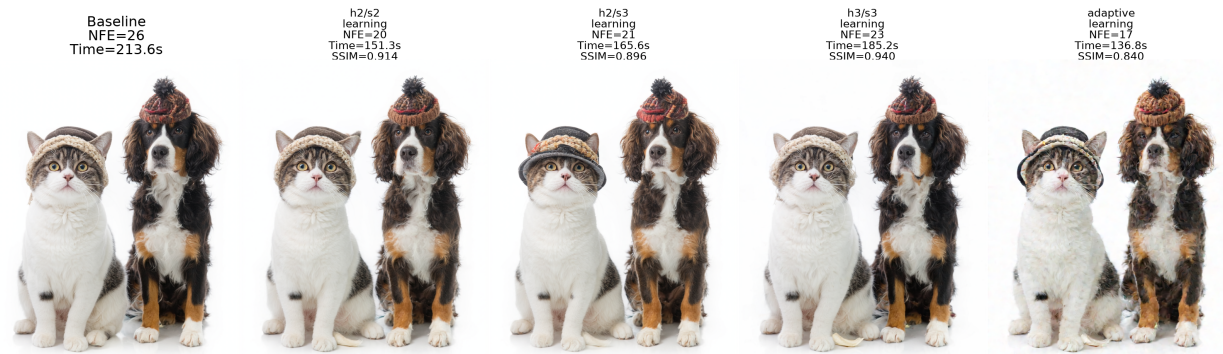

*Figure 4.4b: Wan 2.2 visual comparison. res_2s sampler, two-stage beta+bong_tangent scheduler, 26-step baseline. Best: h3/s5+learning (best by SSIM; SSIM=0.9710, 9.6% time saved).*

Figures 4.4b show visual comparisons for Wan 2.2 (res_2s sampler, two-stage beta+bong_tangent scheduler).

**Learning stabilizer transfers across architectures.** All three models achieve SSIM >= 0.97 using the learning stabilizer, indicating that EMA-based scaling adapts to trajectory curvature regardless of sampler type or scheduler complexity. FLUX.1-dev exhibits the highest efficiency gains (15.9% time saved) with conservative h2/s4+learning because its 20-step baseline allows more aggressive relative skipping. Qwen-Image uses a 25-step euler sampler, yielding excellent quality (SSIM=0.9952) but lower absolute time savings (8.1%) with h2/s5+learning. Wan 2.2's two-stage scheduler (high-noise beta, low-noise bong_tangent) benefits most from h3/s5+learning rather than h2, achieving SSIM=0.9710 with 9.6% time saved; the scheduler transition creates a curvature discontinuity that third-order skip patterns handle more robustly than second-order. Visual inspection finds that SSIM >= 0.95 configurations are near-indistinguishable from baselines with subtle differences across models, validating FSampler's perceptual quality preservation.

## Data Availability

The complete experimental dataset, including all 105 run configurations, raw per-step metrics, analysis scripts, and generated images; code available at: ComfyUI-FSampler, experiment data (including high-resolution images) available at: FSampler_ExperimentData. The experiments directory contains organized subdirectories (fluxr1, qwen1, wan221) with per-experiment metadata, CSV files, and automated analysis reports generated via analyze_experiments.py.

# 5. Conclusion

FSampler addresses the computational bottleneck in diffusion model sampling by introducing fixed-cadence skip patterns with a learning stabilizer (EMA-based scaling). The approach reduces model calls (NFE) while preserving perceptual fidelity by quietly correcting scale on skipped steps. Unlike training-based acceleration, FSampler requires no retraining or architectural changes and drops into existing ComfyUI workflows.

Across FLUX.1-dev, Qwen-Image, and Wan 2.2, representative configurations with SSIM >= 0.95 achieve ~8-22% wall-clock time savings and ~15-25% NFE reduction. For example, on FLUX.1-dev, h2/s4+learning reaches SSIM=0.9818 with 15.9% time saved; a balanced h2/s3+learning variant attains SSIM=0.9533 with

21.6% time saved. In practice, SSIM >= 0.90 corresponds to differences that are often hard to notice; outputs are typically different rather than degraded-the realism and adherence to style are maintained while following an alternate but plausible denoising trajectory (see Section 4).

FSampler is orthogonal to model-level optimizations and precision choices: it combines with distilled models and works with quantized inference, as well as optimized schedulers, compounding efficiency gains at inference. Preliminary use on video pipelines indicates compatibility, though systematic evaluation is needed to quantify temporal coherence and the effect on frame-to-frame consistency.

## Limitations and Broader Impact

FSampler's adaptive gate can achieve ~45-50% NFE reduction but reduces SSIM to ~0.73-0.74 on FLUX.1-dev at those skip rates, indicating a stability boundary for aggressive skipping. The optimal skip pattern (e.g., h2 vs h3) and cadence (s3, s4, s5) vary by sampler architecture and scheduler, requiring per-model selection; guard rails (protected head/tail windows, anchors, and skip caps) are important for stability. Adaptive is user-tunable: the anchor interval forces periodic real calls, max consecutive skips caps back-to-back skips, and explicit step selection (skip indices) lets users specify the exact steps to skip. With per-model tuning, adaptive is especially useful for quick previews of likely final images and fast seed sweeps to find promising candidates before committing conservative skip calls.

The current evaluation emphasizes single-seed comparisons per model and SSIM as the primary perceptual metric. While SSIM >= 0.95 correlates with perceptual similarity in our examples, it does not capture all perceptual dimensions (e.g., semantics, fine texture, or temporal coherence in video). Expanding to multi-prompt benchmarks, human preference studies, and dedicated video evaluations will strengthen generalization claims.

FSampler inherits standard concerns associated with generative model acceleration: faster sampling may facilitate misuse for generating misleading or harmful synthetic content. The method introduces no novel risks beyond existing diffusion model capabilities.

# Bibliography

## Papers

Lu, C., Zhou, Y., Bao, F., Chen, J., Li, C. and Zhu, J. (2022a) 'DPM-Solver: A Fast ODE Solver for Diffusion Probabilistic Model Sampling in Around 10 Steps', *Advances in Neural Information Processing Systems (NeurIPS 2022)*. Available at: https://arxiv.org/abs/2206.00927

Lu, C., Zhou, Y., Bao, F., Chen, J., Li, C. and Zhu, J. (2022b) 'DPM-Solver++: Fast Solver for Guided Sampling of Diffusion Probabilistic Models'. arXiv:2211.01095. Available at: https://arxiv.org/abs/2211.01095

Wang, G., Cai, Y., Li, L., Peng, W. and Su, S. (2025) 'PFDiff: Training-Free Acceleration of Diffusion Models Combining Past and Future Scores', *International Conference on Learning Representations (ICLR 2025)*. Available at: https://arxiv.org/abs/2408.08822

Liu, F., Zhang, S., Wang, X., Wei, Y., Qiu, H., Zhao, Y., Zhang, Y., Ye, Q. and Wan, F. (2025) 'Timestep Embedding Tells: It's Time to Cache for Video Diffusion Model', *IEEE/CVF Conference on Computer Vision and Pattern Recognition (CVPR 2025)*. Available at: https://arxiv.org/abs/2411.19108

Zhao, W., Bai, L., Rao, Y., Zhou, J. and Lu, J. (2023) 'UniPC: A Unified Predictor-Corrector Framework for Fast Sampling of Diffusion Models', *Advances in Neural Information Processing Systems (NeurIPS 2023)*. Available at: https://arxiv.org/abs/2302.04867

Zhang, Q. and Chen, Y. (2022) 'Fast Sampling of Diffusion Models with Exponential Integrator', *International Conference on Learning Representations (ICLR 2023)*. Available at: https://arxiv.org/abs/2204.13902

Song, J., Meng, C. and Ermon, S. (2021) 'Denoising Diffusion Implicit Models', *International Conference on Learning Representations (ICLR 2021)*. Available at: https://arxiv.org/abs/2010.02502

Karras, T., Aittala, M., Aila, T. and Laine, S. (2022) 'Elucidating the Design Space of Diffusion-Based Generative Models', *Advances in Neural Information Processing Systems (NeurIPS 2022)*. Available at: https://arxiv.org/abs/2206.00364

Zhang, Q., Song, J. and Chen, Y. (2023) 'Improved Order Analysis and Design of Exponential Integrator for Diffusion Models Sampling'. arXiv:2308.02157. Available at: https://arxiv.org/abs/2308.02157

## Models Used in Experiments

Black Forest Labs (2024) *FLUX.1 [dev]*. Available at: https://huggingface.co/black-forest-labs/FLUX.1-dev (Accessed: October 2025).

Qwen Team (2024) *Qwen-Image*. Available at: https://huggingface.co/Qwen (Accessed: October 2025).

Wan Team (2024) *Wan 2.2*. Available at: https://huggingface.co/Wan-AI (Accessed: October 2025).